\documentclass{article}

\usepackage[final,nonatbib]{neurips_data_2021}

\usepackage[utf8]{inputenc} 
\usepackage[T1]{fontenc}    
\usepackage{hyperref}
\hypersetup{colorlinks,citecolor=green}
\usepackage{url}            
\usepackage{booktabs}       
\usepackage{amsfonts}       
\usepackage{nicefrac}       
\usepackage{microtype}      
\usepackage{xcolor}         
\usepackage{paralist}
\usepackage{enumitem}
\usepackage{graphicx}
\usepackage{wrapfig}
\newcommand{\R}{\mathbb{R}}

\def\eg{\emph{e.g., }}

\title{ThreeDWorld: A Platform for  \\ Interactive Multi-Modal  Physical Simulation}

\author{
Chuang Gan$^{1}$, Jeremy Schwartz$^{2}$, Seth Alter$^{2}$, Damian Mrowca$^{4}$, Martin Schrimpf$^{2}$, \\  \textbf{James Traer$^{2}$, Julian De Freitas$^{3}$, Jonas Kubilius$^{2}$, Abhishek Bhandwaldar$^{1}$, Nick Haber$^{4}$,} \\ \textbf{ Megumi Sano$^{4}$, Kuno Kim$^{4}$, Elias Wang$^{4}$, Michael Lingelbach$^{4}$, Aidan Curtis$^{2}$, }\\  \textbf{Kevin Feigelis$^{4}$, Daniel M. Bear$^{4}$, Dan Gutfreund$^{1}$, David Cox$^{1}$, Antonio Torralba$^{2}$, }\\\textbf{James J. DiCarlo$^{2}$, Joshua B. Tenenbaum$^{2}$, Josh H. McDermott$^{2}$, Daniel L.K. Yamins$^{4}$} \\ 
\\
 $^1$ MIT-IBM Watson AI Lab,
 $^2$ MIT,
 $^3$ Harvard University,
 $^4$ Stanford University \\\\
 \url{www.threedworld.org}}

\begin{document}

\maketitle

\begin{abstract}
    We introduce ThreeDWorld (TDW), a platform for interactive multi-modal physical simulation. TDW enables simulation of high-fidelity sensory data and physical interactions between mobile agents and objects in rich 3D environments. Unique properties include: real-time near-photo-realistic image rendering; a library of objects and environments, and routines for their customization; generative procedures for efficiently building classes of new environments; high-fidelity audio rendering; realistic physical interactions for a variety of material types, including cloths, liquid, and deformable objects; customizable ``agents” that embody AI agents; and support for human interactions with VR devices. TDW’s API enables multiple agents to interact within a simulation and returns a range of sensor and physics data representing the state of the world. We present initial experiments enabled by TDW in emerging research directions in computer vision, machine learning, and cognitive science, including multi-modal physical scene understanding, physical dynamics predictions, multi-agent interactions, models that ‘learn like a child’, and attention studies in humans and neural networks. 
\end{abstract}

\section{Introduction}

A longstanding goal of research in artificial intelligence is to engineer machine agents that can interact with the world, whether to assist around the house, on a battlefield, or in outer space. Such AI systems must learn to perceive and understand the world around them in physical terms in order to be able to manipulate objects and formulate plans to execute tasks. A major challenge for developing and benchmarking such agents is the logistical difficulty of training an agent. Machine perception systems are typically trained on large data sets that are laboriously annotated by humans, with new tasks often requiring new data sets that are expensive to obtain. And robotic systems for interacting with the world pose a further challenge – training by trial and error in a real-world environment is slow, as every trial occurs in real-time, as well as expensive and potentially dangerous if errors cause damage to the training environment. There is thus growing interest in using simulators to develop and benchmark 
embodied AI and robot learning models~\cite{kolve2017ai2,wu2018building,puig2018virtualhome,savva2019habitat,xia2018gibson,coumans2016pybullet,todorov2012mujoco,xiang2020sapien,beattie2016deepmind}.

World simulators could in principle greatly accelerate the development of AI systems. With virtual agents in a virtual world, training need not be constrained by real-time, and there is no cost to errors (e.g. dropping an object or running into a wall). In addition, by generating scenes synthetically, the researcher gains complete control over data generation, with full access to all generative parameters, including physical quantities such as mass that are not readily apparent to human observers and therefore difficult to label. Machine perceptual systems could thus be trained on tasks that are not well suited to the traditional approach of massively annotated real-world data. A world simulator can also in principle simulate a wide variety of environments, which may be crucial to avoid overfitting.

\begin{figure*}[t]
   \centering
   \includegraphics[width = 1.0\linewidth]{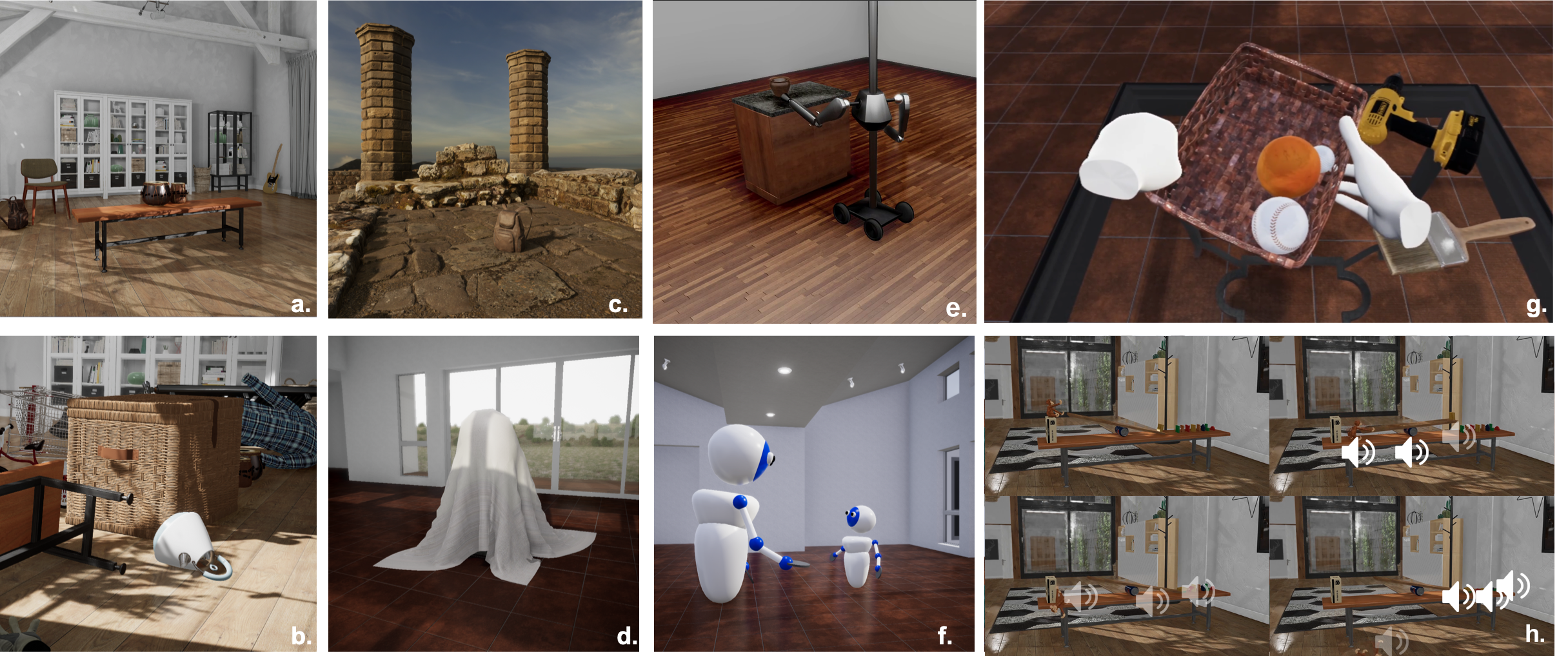}
   \caption{TDW's general, flexible design supports a broad range of use-cases at a high level of multi-modal fidelity: a-c) Indoor and outdoor scene rendering; d) Advanced physics -- cloth draping over a rigid body; e) Robot agent picking up object; f) Multi-agent scene -- "parent" and "baby" avatars interacting; g) Human user interacting with virtual objects in VR; h) Multi-modal scene -- speaker icons show playback locations of synthesized impact sounds.}
\vspace{-5mm}
  \label{fig:teaser}
\end{figure*}

The past several years have seen the introduction of a variety of simulation environments tailored to particular research problems in embodied AI, scene understanding, and physical inference. Simulators have stimulated research in navigation (\eg Habitat \cite{savva2019habitat}, iGibson \cite{xia2020interactive}), robotic manipulation (\eg Sapien \cite{xiang2020sapien}), and embodied learning (\eg AI2Thor \cite{kolve2017ai2}). The impact of these simulators is evident in the many challenges they have enabled in computer vision and robotics. Existing simulators each have various strengths, but because they were often designed with specific use cases in mind, each is also limited in different ways. In principle a system could be trained to see in one simulator, to navigate in another and to manipulate objects in a third. However, switching platforms is costly for the researcher.could generate data with a complete control over data generation

ThreeDWorld (TDW) is a general-purpose virtual world simulation platform that supports multi-modal physical interactions between objects and agents. TDW was designed to accommodate a range of key domains in AI, including perception, interaction, and navigation, with the goal of enabling training in each of these domains within a single simulator. It is differentiated from existing simulation environments by combining high-fidelity rendering for both video and audio, realistic physics, and a single flexible controller.

In this paper, we describe the TDW platform and its key distinguishing features, as well as several example applications that illustrate its use in AI research. These applications include: 1) A learned visual feature representation, trained on a TDW image classification dataset comparable to ImageNet, transferred to fine-grained image classification and object detection tasks; 2) A synthetic dataset of impact sounds generated via TDW’s audio impact synthesis and used to test material and mass classification, using TDW’s ability to handle complex physical collisions and non-rigid deformations; 3) An agent trained to predict physical dynamics in novel settings;  4) Sophisticated multi-agent interactions and social behaviors enabled by TDW’s support for multiple agents; 5) Experiments on attention comparing human observers in VR to a neural network agent. 

A download of TDW's full codebase and documentation is available at: \url{https://github.com/threedworld-mit/tdw}; the code for creating the datasets described below are available at: {\color{blue} \href{ https://github.com/threedworld-mit/tdw/blob/master/Documentation/python/use_cases/single_object.md}{\underline{TDW-Image}}}, {\color{blue} \href{ https://github.com/alters-mit/tdw_sound20k}{\underline{TDW-Sound}}}, and {\color{blue} \href{ https://github.com/alters-mit/tdw_physics}{\underline{TDW-Physics}}}.

\noindent\textbf{Related Simulation Environments}
TDW is distinguished from many other existing simulation environments in the diversity of potential use cases it enables. A summary comparison of TDW’s features to those of existing environments is provided in Table~\ref{tab:comparison}. These environments include AI2-THOR\cite{kolve2017ai2}, HoME\cite{wu2018building}, VirtualHome\cite{puig2018virtualhome}, Habitat\cite{savva2019habitat},  Gibson\cite{xia2018gibson}, iGibson~\cite{xia2020interactive},
Sapien~\cite{xiang2020sapien} PyBullet~\cite{coumans2016pybullet}, MuJuCo~\cite{todorov2012mujoco}, and Deepmind Lab~\cite{beattie2016deepmind}. 

TDW is unique in its support of: a) Real-time near-photorealistic rendering of both indoor and outdoor environments; b) A physics-based model for generating situational sounds from object-object interactions (Fig. \ref{fig:teaser}h); c) Procedural creation of custom environments populated with custom object configurations; d) Realistic interactions between objects, due to the unique combination of high-resolution object geometry and fast-but-accurate high-resolution rigid body physics (denoted ``R$^+$'' in Table \ref{tab:comparison}); e) Complex non-rigid physics, based on the NVIDIA Flex engine; f) A range of user-selectable embodied agent agents; g) A user-extensible model library.

\begin{table}[h]

	\caption{Comparison of TDW's capabilities with those of related virtual simulation frameworks.} 
	\label{tab:comparison}
	\begin{center}
		\begin{tabular}{l|c|c|c|c|c}
		\hline
		  \multicolumn{1}{p{1.0cm}}{\centering Platform}
			 & \multicolumn{1}{|p{1.0cm}}{\centering Scene \\  (I,O)}
			 & \multicolumn{1}{|p{2cm}}{\centering Physics \\  \centering (R/R$^+$,S,C,F)}
			 & \multicolumn{1}{|p{1.25cm}}{\centering Acoustic \\  (E,P)}
			 & \multicolumn{1}{|p{2cm}}{\centering  Interaction \\ (D,A,H)}
			 & \multicolumn{1}{|p{1.25cm}}{\centering  Models \\ (L,E)}\\
			 \hline
			 Deepmind Lab~\cite{beattie2016deepmind} & & & & D, A \\
			 \hline
			 MuJuCo~\cite{todorov2012mujoco} & & R$^+$, C, S & & D, A \\
			 \hline
			 PyBullet~\cite{coumans2016pybullet} & & R$^+$, C, S & E & D, A \\
			 \hline
			 HoME~\cite{wu2018building} & & R & E &  \\ 
			 \hline
			 VirtualHome~\cite{puig2018virtualhome} & I & & & D, A \\  
			 \hline
			 Gibson~\cite{xia2018gibson} & I & & &  \\
			 \hline
			 iGibson~\cite{xia2020interactive} & I & R$^+$ & & D, A ,H & L  \\
			 \hline
			 Sapien~\cite{xiang2020sapien} & I & R$^+$ & & D, A & L  \\
			 \hline
			 Habitat~\cite{savva2019habitat} & I & & E&  \\
			 \hline
			 AI2-THOR~\cite{kolve2017ai2} & I & R  &  &  D & L\\
			 \hline
			 ThreeDWorld & I, O & R$^+$, C, S, F & E, P & D, A, H & L, E\\
			 \hline
		\end{tabular}
	\end{center}
\end{table}
\noindent\textbf{Summary:}  Table~\ref{tab:comparison} shows TDW differs from these  frameworks in its support for different types of:
\setdefaultleftmargin{1em}{1em}{}{}{}{}
\begin{compactitem}
	\item Photorealistic scenes: indoor (I) and outdoor (O)
	\item Physics simulation: just rigid body (R) or improved fast-but-accurate rigid body (R$^+$), soft body (S), cloth (C) and fluids (F)
	\item Acoustic simulation: environmental (E) and physics-based (P)
	\item User interaction: direct API-based (D), agent-based (A) and human-centric using VR (H)
	\item Model library support: built-in (L) and user-extensible (E)
\end{compactitem}

\section{ThreeDWorld Platform}

\subsection{Design Principles and System Overview}

\textbf{Design Principles.}  Our core contribution is to integrate several existing real-time advanced physics engines into a framework that can also produce high-quality visual and auditory renderings. In making this integration, we followed three design principles:
\setdefaultleftmargin{1em}{1em}{}{}{}{}
\begin{compactitem}
\item The integration should be flexible. That is, users should be able to easily set up a wide variety of physical scenarios, placing any type of object at any location in any state, with controllable physical parameters. This enables researcher to create physics-related benchmarks with highly variable situations while also being able to generate near-photorealistic renderings of those situations.

\item The physics engines should cover a wide variety of object interactions. We achieve this aim by seamlessly integrating PhysX (a good rigid-body simulator) and Nvdia Flex (a state-of-the-art multi-material simulator for non-rigid and rigid-non-rigid interactions).

\item There should be a large library of high-quality assets with accurate physical descriptors as well as realistic rigid and non-rigid material types, to allow users to take advantage of the power of the physics engines and easily be able to produce interesting and useful physical scenes. 
\end{compactitem}

\textbf{System Overview.} The TDW simulation consists of two basic components: (i) the \textbf{Build}, a compiled executable running on the Unity3D Engine, which is responsible for image rendering, audio synthesis and  physics simulations; and (ii) the \textbf{Controller}, an external Python interface to communicate with the build. Users can define their own tasks through it, using an API comprising over 200 commands.
Running a simulation follows a cycle in which:
1) The controller sends \textbf{commands} to the build; 2) The build executes those commands and sends \textbf{simulation output data} back to the controller. Unlike other simulation platforms, TDW’s API commands can be combined into lists and sent to the build within a single time step, allowing the simulation of arbitrarily complex behavior. Researchers can use this core API as a foundation on which to build higher-level, application-specific API "layers" that dramatically reduce development time and enable widely divergent use cases.

\subsection{Photo-realistic Rendering}

TDW uses Unity's underlying game-engine technology for image rendering, adding a custom lighting approach to achieve near-photorealistic rendering quality for both indoor and outdoor scenes.

\noindent\textbf{Lighting Model.} TDW uses two types of lighting; a single light source simulates direct light coming from the sun, while indirect environment lighting comes from “skyboxes” that utilize High Dynamic Range (HDRI) images. For details, see Fig~\ref{fig:teaser}(a-c) and the Supplement. Additional post-processing is applied to the virtual camera including exposure compensation, tone mapping and dynamic depth-of-field ({\color{blue} \href{https://bit.ly/2UeSggG}{\underline{examples}}}). 

\noindent\textbf{3D Model Library.}
To maximize control over image quality we have created a library of 3D model “assets” optimized from high-resolution 3D models. Using Physically-Based Rendering (PBR) materials, these models respond to light in a physically correct manner. The library contains around 2500 objects spanning 200 categories organized by Wordnet synset, including furniture, appliances, animals, vehicles, and toys etc.  Our material library contains over 500 materials across 10 categories, many scanned from real world materials.

\noindent\textbf{Procedural Generation of New Environments.}
In TDW, a run-time virtual world, or ``scene”, is created using our 3D model library assets. Environment models (interior or exterior) are populated with object models in various ways, from completely procedural (i.e. rule-based) to thematically organized (i.e. explicitly scripted). TDW places no restrictions on which models can be used with which environments, which allows for unlimited numbers and types of scene configurations.

\subsection{High-fidelity Audio Rendering}
\label{sec:audio}
Multi-modal rendering is an unique aspect of TDW, and our audio engine provides both physics-driven impact sound generation, and reverberation and spatialized sound simulation.

\noindent\textbf{Generation of Impact Sounds.}
TDW's includes PyImpact, a Python library that uses modal synthesis to generate impact sounds \cite{traer2019impacts}. PyImpact uses information about physical events such as material types, as well as velocities, normal vectors and masses of colliding objects to synthesize sounds that are played at the time of impact {(\color{blue} \href{https://bit.ly/3cAh2y6}{\underline{examples}})}.  This “round-trip” process is real-time. Synthesis is currently being extended to encompass scraping and rolling sounds \cite{agarwal2021scrapingrolling}.


\noindent\textbf{Environmental Audio and Reverberation.}
For sounds placed within interior environments, TDW uses a combination of Unity's built-in audio and Resonance Audio's 3D spatialization to provide real-time audio propagation, high-quality simulated reverberation and directional cues via head-related transfer functions. Sounds are attenuated by distance and can be occluded by objects or environment geometry. Reverberation automatically varies with the geometry of the space, the virtual materials applied to walls, floor and ceiling, and the percentage of room volume occupied by solid objects (\eg furniture). 

\subsection{Physical Simulation}


 \begin{wrapfigure}{r}{0.4\linewidth}
\vspace{-4mm}

\small\centering
\includegraphics[width=0.94\linewidth]{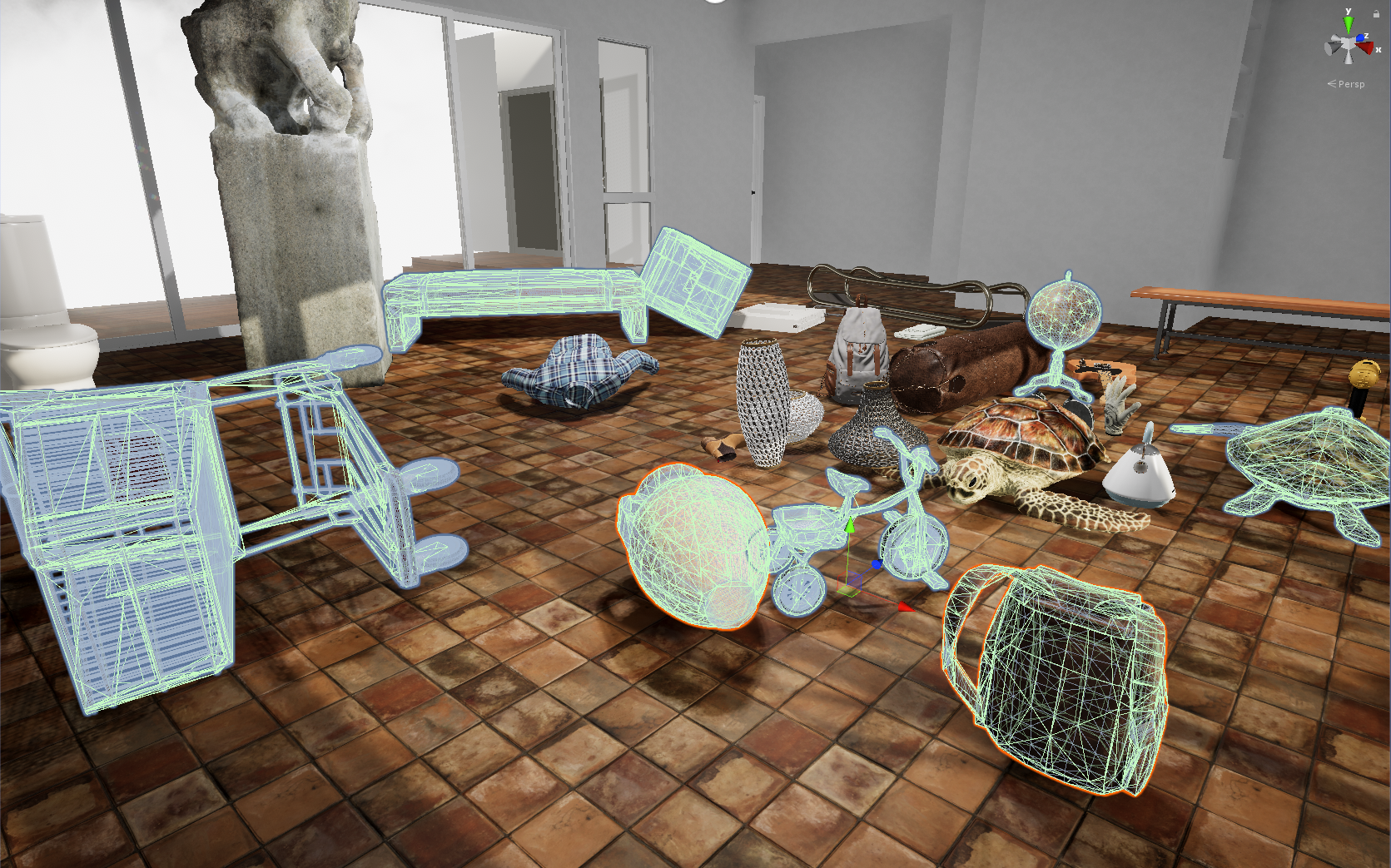}
\vspace{-2mm}
\caption{Green outlines around objects indicate auto-computed convex colliders for fast but accurate rigid-body physics.}
\vspace{-5mm}
\label{fig:rightbody}
\end{wrapfigure}

In TDW, object behavior and interactions are handled by a physics engine. TDW now integrates two physics engines, supporting both rigid-body physics and more advanced soft-body, cloth and fluid simulations.

\textbf{Rigid-body physics.}
Unity’s rigid body physics engine (PhysX) handles basic physics behavior involving collisions between rigid bodies. To achieve accurate but efficient collisions, we use the powerful V-HACD algorithm~\cite{mamou2009simple} to compute ``form-fitting'' convex hull colliders around each library object’s mesh, used to simplify collision calculations (see Figure 2).  In addition, an object's mass is automatically calculated from its volume and material density upon import. 
However, using API commands it is also possible to dynamically adjust mass or friction, as well as visual material appearance, on a per-object basis enabling potential disconnection of visual appearance from physical behavior (e.g. objects that look like concrete but bounce like rubber).

\textbf{Advanced Physics Simulations.}
TDW's second physics engine – Nvidia Flex -- uses a particle-based representation to manage collisions between different object types. TDW supports rigid body, soft body (deformable), cloth and fluid simulations Figure~\ref{fig:teaser}(d). This unified representation helps machine learning models use underlying physics and rendered images to learn a physical and visual representation of the world through interactions with objects in the world.


\subsection{Interactions and Agents}
TDW provides three paradigms for interacting with 3D objects: 1) Direct control of object behavior using API commands. 2) Indirect control through an embodiment of an AI agent. 3) Direct interaction by a human user, in virtual reality (VR).

\noindent\textbf{Direct Control.}
Default object behavior in TDW is completely physics-based via commands in the API; there is no scripted animation of any kind. Using physics-based commands, users can move an object by applying an impulse force of a given magnitude and direction.

\noindent\textbf{Agents.}
The embodiment of AI agents come in several types: 
\setdefaultleftmargin{1em}{1em}{}{}{}{}
\begin{compactitem}
	\item Disembodied cameras for generating first-person rendered images, segmentation and depth maps.
	\item Basic embodied agents whose avatars are geometric primitives such as spheres or capsules that can move around the environment and are often used for algorithm prototyping. 
	\item More complex embodied avatars with user-deﬁned physical structures and associated physically-mapped action spaces. For example, TDW’s Magnebot is a complex robotic body, fully physics-driven with articulated arms terminating in 9-DOF end-effectors (Fig. \ref{fig:teaser}e). By using commands from its high-level API such as \textbf{reach\_for(target position)} and \textbf{grasp(target object)}, Magnebot can be made to open boxes or pick up and place objects. In addition, as a first step towards sim2real transfer, researchers can also import standard URDF robot specification files into TDW and use actual robot types such as Fetch, Sawyer or Baxter as embodied agents. 
\end{compactitem}

Agents can move around the environment while responding to physics, using their physics-driven articulation capabilities to change object or scene states, or can interact with other agents within a scene (Fig. \ref{fig:teaser}f). 

\noindent\textbf{Human Interactions with VR devices.} TDW also supports users interacting directly with 3D objects using VR. Users see a 3D representation of their hands that tracks the actions of their own hands (Fig. \ref{fig:teaser}g). Using API commands, objects are made ``graspable" such that any collision between object and virtual hands allows the user to pick it up, place it or throw it ({\color{blue} \href{https://bit.ly/3cAtH46}{\underline{example}}}).  This functionality enables the collection of human behavior data, and allows humans to interact with avatars.

\section{Example Applications}
\label{sec:tasks}

\subsection{Visual and Sound Recognition Transfer}
We quantitatively examine how well feature representations learned using TDW-generated images and audio data transfer to real world scenarios. 

\textbf{Visual recognition transfer} We generated a TDW image classification dataset  comparable in size to ImageNet; 1.3M images were generated by randomly placing one of TDW's 2,000 object models in an environment with random conditions (weather, time of day)
and taking a snapshot while pointing the randomly positioned virtual camera at the object (\textbf{ Details in Supplement}).

\begin{table}[http]
\small

 \caption{Visual representations transfer for  fine-grained image classifications.}

 \label{tab:vision}
 \begin{center}
  \begin{tabular}{l|ccccccc|c}
 \hline
   Dataset &  Aircraft & Bird & Car & Cub & Dog & Flower & Food & Mean\\
 \hline
   ImageNet & 0.74 & 0.70 & 0.86 & 0.72 & 0.72 & 0.92 & 0.83 & 0.78\\
   SceneNet & 0.06 & 0.43 & 0.30 & 0.27 & 0.38 & 0.62 & 0.77 & 0.40\\
   AI2-THOR & 0.57 & 0.59 & 0.69 & 0.56 & 0.56 & 0.62 & 0.79 & 0.63\\
   
 \hline
   TDW & 0.73 & 0.69 & 0.86 & 0.7 & 0.67 & 0.89 & 0.81 & 0.76\\

 \hline
     \end{tabular}
\end{center}
\end{table}
We pre-trained four ResNet-50 models \cite{he2016deep} on ImageNet~\cite{deng2009imagenet}, SceneNet~\cite{handa2016understanding}, AI2-Thor~\cite{kolve2017ai2} and the TDW-image dataset respectively. We directly downloaded images of ImageNet~\cite{deng2009imagenet} and SceneNet~\cite{handa2016understanding} for model trainings. For a fair comparison, we also created an AI2-THOR dataset with 1.3M images using a controller that captured random images in a scene and classified its segmentation masks from ImageNet synset IDs. We then evaluated the learned representations by fine-tuning on downstream fine-grained image classification tasks using Aircraft~\cite{maji2013fine}, Birds~\cite{van2015building}, CUB~\cite{wah2011caltech}, Cars~\cite{krause20133d}, Dogs~\cite{khosla2011novel}, Flowers~\cite{nilsback2006visual}, and Food datasets~\cite{bossard2014food}.  We used a ResNet-5- network architecture as a backbone for all the visual perception transfer experiments. For the pre-training, we set the initial learning rate as 0.1 with cosine decay and trained for 100 epochs. We then took the pre-trained weights as initialization and fine-tuned on fine-grained image recognition tasks, using an initial learning rate of 0.01 with cosine decay and training for 10 epochs on the fine-grained image recognition datasets. Table~\ref{tab:vision} shows that the feature representations learned from TDW-generated images are substantially better than the ones learned from SceneNet~\cite{handa2016understanding} or AI2-Thor~\cite{kolve2017ai2}, and have begun to approach the quality of those learned from ImageNet. These experiments suggest that though significant work remains, TDW has taken meaningful steps towards mimicking the use of large-scale real-world datasets in model pre-training. Using a larger transformer architecture~\cite{dosovitskiy2020image} with more TDW-generated images might further close the gap with Imagenet pre-trained models on object recognition tasks. We have open-sourced the full image generation codebase to support future research in directions such as this.

\textbf{Sound recognition transfer} We also created an audio dataset to test material classification from impact sounds. We recorded 300 sound clips of 5 different materials (cardboard, wood, metal, ceramic, and glass; between 4 and 15 different objects for each material) each struck by a selection of pellets (of wood, plastic, metal; of a range of sizes for each material) dropped from a range of heights between 2 and 75cm. The pellets themselves resonated negligible sound compared to the objects but because each pellet preferentially excited different resonant modes, the impact sounds depend upon the mass and material of the pellets, and the location and force of impact, as well as the material, shape, and size of the resonant objects~\cite{traer2019impacts} ({\color{blue} \href{https://bit.ly/2A20im6}{\underline{more video examples}}}). 

Given the variability in other factors, material classification from this dataset is nontrivial. We trained material classification models on simulated audio from both TDW and the sound-20K dataset\cite{zhang2017shape}. We tested their ability to classify object material from the real-world audio.  We  converted the raw audio waveform to a sound spectrogram representation and fed them to a VGG-16 pre-trained on AudioSet \cite{AudioSet}. For the material classification training, we set the initial learning rate as 0.01 with cosine decay and trained for 50 epochs.  As shown in Table~\ref{tab:sound}, the model trained on the TDW audio dataset achieves more than 30\% better accuracy gains than that trained on the Sound20k dataset. This improvement is plausibly because TDW produces a more diverse range of sounds than Sound20K and prevents the network overfitting to specific features of the synthetic audio set. 

\begin{table}[!http]
\begin{minipage}[t]{.4\linewidth}
\centering
 \vspace{-2mm}  
\caption{Sound perception transfer on material recognition.}

  \begin{tabular}{l|c}
  \hline
   Dataset & Accuracy \\
   \hline

    Sound-20K &0.34 \\
    TDW & \textbf{0.66}  \\
   \hline
  \end{tabular}

  \label{tab:sound}

\end{minipage}
\hfill
\begin{minipage}[t]{.55\linewidth}
 \vspace{-2mm}  
 \caption{Comparison of the multi-modal physical scene understanding on material and mass classification.}
 \label{tab:multi-model}
  \centering
  \begin{tabular}{l|c|c}
 \hline
   Method & Material  &  Mass \\
 \hline
   Vision only & 0.72  &  0.42\\
   Audio only & 0.92 &0.78\\
 \hline
   Vision + Audio &  \textbf{0.96} & \textbf{0.83}\\
 \hline
 \end{tabular}
\end{minipage}

\end{table}
\textbf{Multi-modal physical scene understanding} We used the TDW graphics engine, physics simulation and the sound synthesis technique described in Sec~\ref{sec:audio} to generate videos and impact sounds of objects dropped on flat surfaces (table tops and benches). The surfaces were rendered to have the visual appearance of one of 5 materials. The high degree of variation over object and material appearance, as well as physical properties such as trajectories and elasticity, prevents the network from memorizing features (i.e. that objects bounce more on metal than cardboard). The training and test sets had the same material and mass class categories.  However, the test-set videos contained objects, tables, motion patterns, and impact sounds that were different from any video in the training set.  Across all videos, the identity, size, initial location, and initial angular momentum of the dropped object were randomized to ensure every video had a unique pattern of motion and bounces. The shape, size, and orientation of the table were randomized, as were the surface texture renderings (e.g., a wooden table could be rendered as "cedar," "pine," "oak," "teak," etc.), to ensure every table appearance was unique. PyImpact uses a random sampling of resonant modes to create an impact sound, such that the impacts in every video had a unique spectro-temporal structure. 

 For the vision-only baseline, we  extracted visual features from each video frame using a ResNet-18 pre-trained on ImageNet, applying an average pooling over 25 video frames to arrive a 2048-d feature vector. For the audio-only baseline, we converted the raw audio waveforms to sound spectrograms and provided them as input for a VGG-16 pre-trained on AudioSet. Each audio-clip was then represented as a 4096-d feature vector. We then took the visual-only features, sound-only features, and the concatenation of visual and sound feature as input to a 2-layer MLP classifier trained for material and mass classification. The results (Table~\ref{tab:multi-model}) show that audio is more diagnostic than video for both classification tasks, but that the best performance requires audiovisual (i.e. multi-modal) information, underscoring the utility of realistic multi-modal rendering.

\subsection{Training and Testing Physical Dynamics Understanding}
Differentiable forward predictors that mimic human-level intuitive physical understanding have emerged as being of importance for enabling deep-learning based approaches to model-based planning and control applications~\cite{Lerer2016Learning,Battaglia2013Simulation,Mottaghi2016What,Fragkiadaki2016Learning,Battaglia2016Interaction,Chang2017compositional,Agrawal2016Learning,Shao2014Imagining,finn2016unsupervised,fire2016learning,pearl2009causality,YeTian_physicsECCV2018}.  
While traditional physics engines constructed for computer graphics (such as PhysX and Flex) have made great strides, such routines are often hard-wired, and thus both hard to apply to novel physical situations encountered by real-world robots, and challenging to integrate as components of larger learnable systems. 
Creating end-to-end differentiable neural networks for intuitive physics prediction is thus an important area of research.
However, the quality and scalability of learned physics predictors has been limited, in part by the availability of effective training data. 
This area has thus afforded a compelling use case for TDW, highlighting its advanced physical simulation capabilities.

\begin{figure}[http]
\begin{center}

\includegraphics[width = 1.0\linewidth]{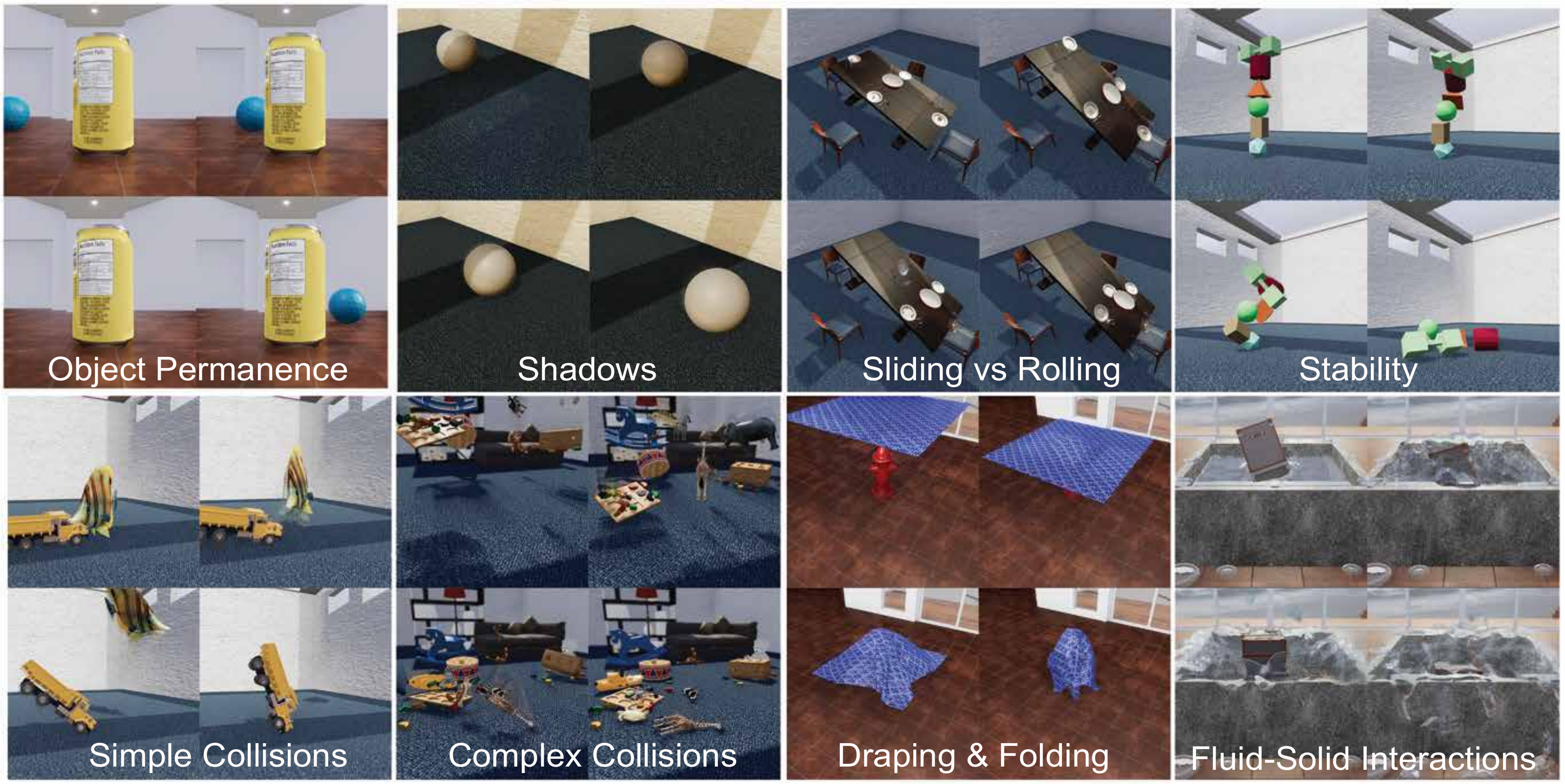}
\end{center}

\caption{\textbf{Advanced Physical Understanding Benchmark.}  Scenarios for training and evaluating advanced physical understanding in end-to-end differentiable physics predictors. These are part of a benchmark dataset that will be released along with TDW. Each panel of four images is in order of top-left, top-right, bottom-left, bottom-right ( {\color{blue} \href{https://bit.ly/3eQX6Za}{\underline{more video examples}}}).
}

\label{fig:physics_benchmark}
\end{figure}

\textbf{Advanced Physical Prediction Benchmark} Using the TDW platform, we have created a comprehensive Pysion benchmark for training and evaluation of physically-realistic forward prediction algorithms~\cite{bear2021physion}. 
This dataset contains a large and varied collection of physical scene trajectories, including all data from visual, depth, audio, and force sensors, high-level semantic label information for each frame, as well as latent generative parameters and code controllers for all situations.
This dataset goes well beyond existing related benchmarks, such as IntPhys~\cite{riochet2018intphys}, providing scenarios with large numbers of complex real-world object geometries, photo-realistic textures, as well as a variety of rigid, soft-body, cloth, and fluid materials. 
Example scenarios from this dataset are seen in Fig \ref{fig:physics_benchmark} are grouped into subsets highlighting important issues in physical scene understanding, including: 
\begin{compactitem}
\item Object Permanence: Object Permanence is a core feature of human intuitive physics~\cite{spelke1990principles}, and agents must learn that objects continue to exist when out of sight.

\item Shadows: TDW’s lighting models allows agents to distinguish both object intrinsic properties (e.g. reflectance, texture) and extrinsic ones (what color it appears), which is key to understanding that appearance can change depending on context, while underlying physical properties do not.

\item Sliding vs Rolling: Predicting the difference between an object rolling or sliding – an easy task for adult humans -- requires a sophisticated mental model of physics. Agents must understand how object geometry affects motion, plus some rudimentary aspects of friction.

\item Stability: Most real-world tasks involve some understanding of object stability and balance. Unlike simulation frameworks where object interactions have predetermined stable outcomes, using TDW agents can learn to understand how geometry and mass distribution are affected by gravity.

\item Simple Collisions: Agents must understand how momentum and geometry affects collisions to know that what happens when objects come into contact affects how we interact with them.

\item Complex Collisions: Momentum and high resolution object geometry help agents understand that large surfaces, like objects, can take part in collisions but are unlikely to move.

\item Draping \& Folding: By modeling how cloth and rigid bodies behave differently, TDW allows agents to learn that soft materials are manipulated into different forms depending on what they are in contact with.

\item Submerging: Fluid behavior is different than solid object behavior, and interactions where fluid takes on the shape of a container and objects displace fluid are important for many real-world tasks.
\end{compactitem}

\begin{figure}[http]
\begin{center}

\includegraphics[width=1.0\textwidth]{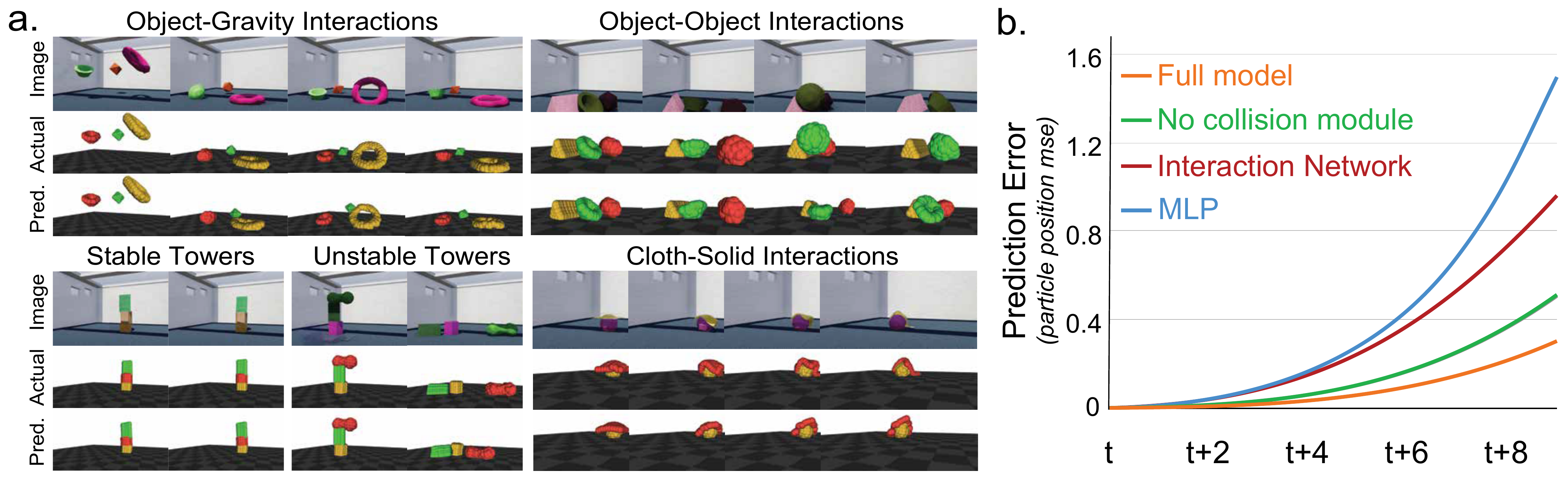}
\end{center}

\caption{\textbf{Training a Learnable Physics Simulator.}  \textbf{(a)} Examples of prediction rollouts for a variety of physical scenarios. \textbf{b)} Quantative evaluations of physical predictions over time for HRN compared to no-collision ablation (green), Interaction Network~\cite{Battaglia2016Interaction} (red), and simple MLP (blue). } 

\label{fig:physics_algs}
\end{figure}

\textbf{Training a Learnable Intuitive Physics Simulator} The Hierarchical Relation Network (HRN) is a recently-published end-to-end differentiable neural network based on hierarchical graph convolution, that learns to predict physical dynamics in this representation~\cite{mrowca2018flexible}. 
The HRN relies on a hierarchical part-based object representation that covers a wide variety of types of three-dimensional objects, including both arbitrary rigid geometrical shapes, deformable materials, cloth, and fluids.
Here, we train the HRN on large-scale physical data generated by TDW, as a proof of concept for TDW's physical simulation capabilities. 
Building on the HRN, we also introduce a new Dynamic Recurrent HRN (DRHRN) (\textbf{Network Details in Supplement}).  that achieves improved physical prediction results that take advantage of the additional power of the TDW dataset generation process.

\begin{table}[http]
\begin{center}
\vspace{-2mm}
\caption{\textbf{Improved Physical Prediction Models.} We measure the global (G) and local (L) position MSE and show qualitative predictions of our \textbf{DRHRN} model at \emph{40} time steps in the future on Lift, Slide, Collide, Stack and Cloth data. $|N|$ is the number of objects in the scene.}
\label{fig:drhrn}
\begin{tabular}{|@{\hskip2pt}c@{\hskip2pt}||c|c|c|c|c|c|c|c|c|c|c|}

\hline
[G] \textbf{$\mathbf{\times 10^{ - 1}}$ }
& \multicolumn{2}{c|}{\textbf{Lift |3|}} &
\multicolumn{2}{c|}{\textbf{Slide |3|}} &
\multicolumn{2}{c|}{\textbf{Collide |3|}} & \multicolumn{2}{c|}{\textbf{Stack |3|}} &
\multicolumn{2}{c|}{\textbf{Cloth |2|}} \\
\cline{2-11}
[L] \textbf{$\mathbf{\times 10^{ - 2}}$ } & 
\ \ \ G \ \ \ & \ \ \ L \ \ \ & 
\ \ \ G \ \ \ & \ \ \ L \ \ \ & 
\ \ \ G \ \ \ & \ \ \ L \ \ \ & 
\ \ \ G \ \ \ & \ \ \ L \ \ \ & 
\ \ \ G \ \ \ & \ \ \ L \ \ \ \\

\textbf{HRN}~\cite{mrowca2018flexible} & 3.27 & 4.18 & 2.04 & 3.89 & 4.08 & 4.34 & 3.50 & 2.94 & 1.33 & 2.22 \\
\hline
\textbf{DPI}\cite{li2018learning} & 3.37 & 4.98 & 3.25 & 3.42 & 4.28 & 4.13 & 3.16 & 2.12 & 0.42 & 0.97 \\
\textbf{DRHRN} & \textbf{1.86} & \textbf{2.45} & \textbf{1.29} & \textbf{2.36} & \textbf{2.45} & \textbf{2.98} & \textbf{1.90} & \textbf{1.83} & \textbf{0.24} & \textbf{0.64} \\
\hline

\end{tabular}

\includegraphics[width=1.0\textwidth]{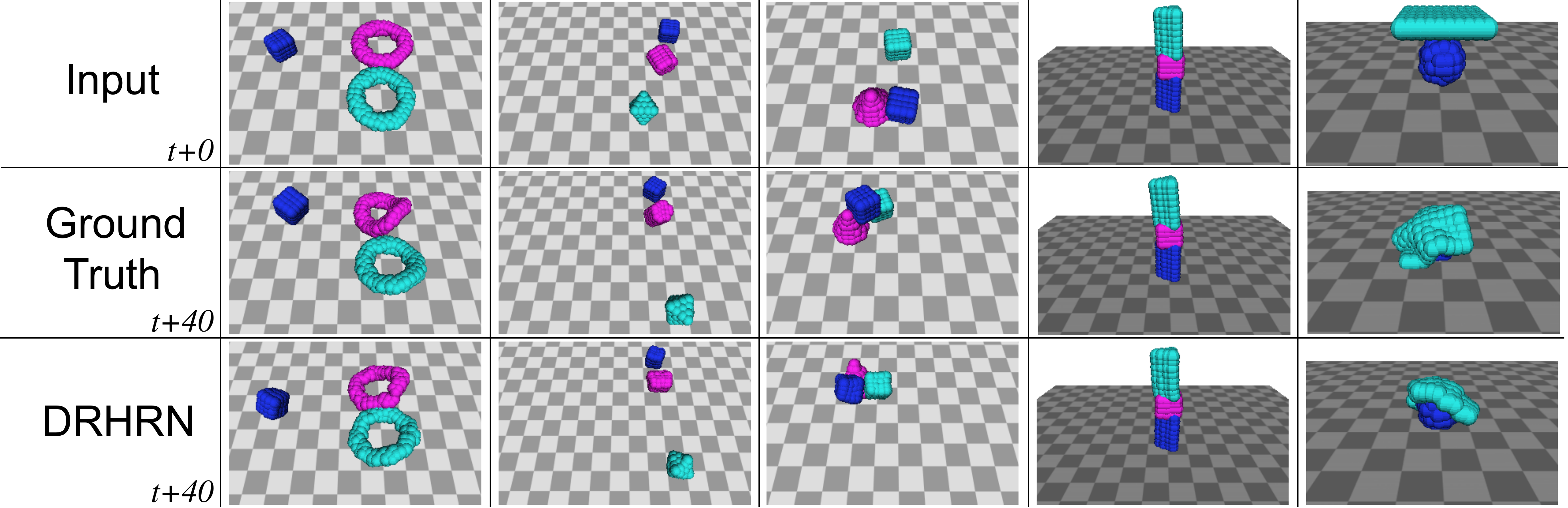}
\vspace{-3mm}

\end{center}
\end{table}

\noindent\textbf{Experimental settings} To evaluate HRN and DRHRN accuracy and generalization, we utilize a subset of the scenarios in the advanced physical understanding benchmark.
We use objects of different shapes (\emph{bowl}, \emph{cone}, \emph{cube}, \emph{dumbbell}, \emph{octahedron}, \emph{pentagon}, \emph{plane}, \emph{platonic}, \emph{prism}, \emph{ring}, \emph{sphere}) and materials (\emph{cloth}, \emph{rigid}, \emph{soft}) to construct the following scenarios: (1) A \textbf{lift subset}, in which objects are lifted and fall back on the ground. (2) A \textbf{slide subset}, in which objects are pushed horizontally on a surface under friction. (3) A \textbf{collide subset}, in which objects are collided with each other. (4) A \textbf{stack subset}, in which objects are (un)stably stacked on top of each other. And (5) a \textbf{cloth subset}, in which a cloth is either dropped on one object or placed underneath and lifted up. Three objects are placed in the first four scenarios, as at least three objects are needed to learn indirect object interactions (e.g. stacking).
Each subset consists of 256-frame trajectories, 350 for training (\textasciitilde90,000 states) and 40 for testing (\textasciitilde10,000 states). 

Given two initial states, each model is trained to predict the next future state(s) at 50 ms intervals. We train models \emph{on all train subsets at once} and evaluate on test subsets separately. We measure the mean-square-error (MSE) between predicted and true particle positions in global and local object coordinates. Global MSE quantifies object position correctness. Local MSE assesses how accurately the object shape is predicted. We evaluate predictions 40 frames into the future. For a better visualization of training and test setups, please follow this {\color{blue} \href{https://drive.google.com/drive/folders/1RCsEDL1LGHyFjhskIjmk1JNZzP96FP_a?usp=sharing}{\underline{video link}}}.

\noindent\textbf{Prediction Results} 
We first replicate results comparing the HRN against simpler physical prediction baselines. As in the original work, we find that HRN outperforms baseline models without collision-detection or flexible hierarchical scene description (Fig. \ref{fig:physics_algs}).  We then compare DRHRN against strong deterministic physics prediction baselines, including HRN as above, and DPI \cite{li2018learning}, which uses a different hierarchical message passing order and a hard coded rigid shape preservation constraint. We re-implement both baselines in Tensorflow for direct comparison.  Table \ref{fig:drhrn} presents results of the DRHRN comparison. DRHRN clearly outperforms HRN and DPI on all scenarios. It achieves a lower local MSE, indicating better shape preservation which we can indeed observe in the images. All predictions look physically plausible without unnatural deformations  ({\color{blue} \href{https://drive.google.com/drive/folders/1RCsEDL1LGHyFjhskIjmk1JNZzP96FP_a?usp=sharing}{\underline{more video results}}}).

\subsection{Social Agents and Virtual Reality}
\label{subsec:vr_experiment}
Social interactions are a critical aspect of human life, but are an area where current approaches in AI and robotics are especially limited. AI agents that model and mimic social behavior, and that learn efficiently from social interactions, are thus an important area for cutting-edge technical development.

\begin{figure}[http]
	\centering
	\includegraphics[height=0.34\linewidth]{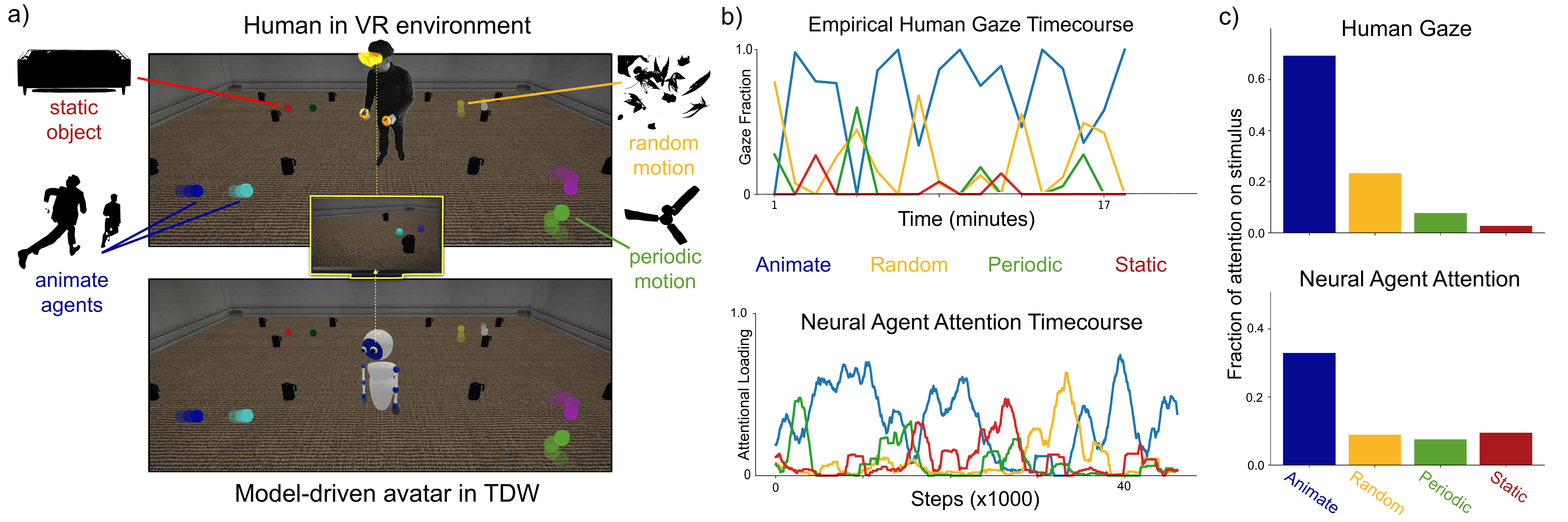}
	\caption{\textbf{Multi-Agent and VR Capabilities.} \textbf{a)} Illustration of TDW’s VR capabilities in an experiment measuring spontaneous patterns of attention to agents executing spatiotemporal kinematics typical of real-world inanimate and animate agents. By design, the stimuli are devoid of surface features, so that both humans and intrinsically-motivated neural network agents must discover which agents are interesting and thus worth paying attention to, based on the behavior of the actor agents. Example timecourses (panel \textbf{b}) and aggregate attention (panel \textbf{c}) for different agents, from humans over real time, and from intrinsically-motivated neural network agents over learning time.}

	\label{fig:VR_scene1}
\end{figure}

\textbf{Task Definition}
Using the flexibility of TDW's multi-agent API, we have created implementations of a variety of multi-agent interactive settings (Fig. \ref{fig:teaser}f). 
These include scenarios in which an ``observer” agent is placed in a room with multiple inanimate objects, together with several differentially-controlled ``actor” agents (Fig. \ref{fig:VR_scene1}a). The actor agents are controlled by either hard-coded or interactive policies implementing behaviors such as object manipulation, chasing and hiding, and motion imitation. Human observers in this setting are simply asked to look at whatever they want, whereas our virtual observer seeks to maximize its ability to predict the behaviors of the actors in this same display, allocating its attention based on a metric of ``progress curiosity''~\cite{baranes2013active} that seeks to estimate which observations are most likely to increase the observer's ability to make actor predictions. The main question is whether this form of curiosity-driven learning naturally gives rise to patterns of attention that mirror how humans allocate attention as they explore this same scene for the first time during the experiment.

\textbf{Experiments}
Intriguingly, in recent work, these socially-curious agents have been shown to outperform a variety of existing alternative curiosity metrics in producing better predictions, both in terms of final performance and substantially reducing the sample complexity required to learn actor behavior patterns~\cite{kim2020progress}. 
The VR integration in TDW enables humans to directly observe and manipulate objects in responsive virtual environments. Fig.~\ref{fig:VR_scene1} illustrates an experiment investigating the patterns of attention that human observers exhibit in an environment with multiple animate agents and static objects~\cite{johnson2003detecting, frankenhuis2013infants}. Observers wear a GPU-powered Oculus Rift S, while watching a virtual display containing multiple robots. Head movements from the Oculus are mapped to a sensor camera within TDW, and camera images are paired with meta-data about the image-segmented objects, in order to determine which set of robots people are gazing at.
Interestingly, the socially-curious neural network agents produce an aggregate attentional gaze pattern that is quite similar to that of human adults measured in the VR environment (Fig.~\ref{fig:VR_scene1}b), arising from the agent’s discovery of the inherent relative ``interestingness” of animacy, without building it in to the network architecture~\cite{kim2020progress}.  These results are just one illustration of TDW’s extensive VR capabilities in bridging AI and human behaviors.

\section{Future Directions}
We are actively working to develop new capabilities for robotic systems integration and articulatable object interaction for higher-level task planning and execution.  
\noindent\textbf{\textbf{Articulatable Objects.}}
Currently only a small number of TDW objects are modifiable by user interaction, and we are actively expanding the number of library models that support such behaviors, including containers with lids that open, chests with removable drawers and doors with functional handles.
\noindent\textbf{\textbf{Humanoid Agents.}}
Interacting with actionable objects or performing ﬁne-motor control tasks such as solving a jigsaw puzzle requires agents with a fully articulated body and hands. We plan to develop a set of humanoid agent types that fulﬁll these requirements, with body movement driven by motion capture data and a separate gesture control system for fine motor control of hand and finger articulation. 
\noindent\textbf{\textbf{Robotic Systems Integration.}}
Building on the modular API layering approach, we envision developing additional ``ultra-high-level” API layers to address specific physical interaction scenarios. We are also exploring creating a PyBullet \cite{coumans2016pybullet} ``wrapper” that would allow replicating physics behaviors between systems by converting PyBullet API commands into comparable commands in TDW.

\begin{ack}
This work was supported by MIT-IBM Watson AI Lab and its member company Nexplore, ONR MURI, DARPA Machine Common Sense program, ONR (N00014-18-1-2847), Mitsubishi Electric, and NSF Grant BCS-192150.
\end{ack}

\bibliographystyle{plain}
\bibliography{egbib}

\newpage

\appendix

\begin{center}
{
\LARGE \textbf{Supplementary Material}
}
\end{center}

In this supplement, we start by discussing the broader impact of TDW. Section~\ref{sec: dataset} discusses implementation details of the TDW image dataset, and explains the setup of each scenario in the Advanced Physical Prediction Benchmark dataset described in the paper. Section~\ref{dynamics} introduces the details of training the HRN~\cite{mrowca2018flexible} and new proposed DRHRN for physical dynamics predictions. We then elaborate on the lighting model used in TDW, and in Section~\ref{sec:related} discuss in more detail how TDW compares to other simulation environments. Lastly, Section~\ref{sec:system} provides a detailed overview of TDW's system architecture, API, benchmarks and code examples showing both back-end and front-end functionality.

\section{Broader Impact}
\label{impact}
As we have illustrated, TDW is a completely general and flexible simulation platform, and as such can benefit research that sits at the intersection of neuroscience, cognitive science, psychology, engineering and machine learning / AI. We feel the broad scope of the platform will support research into understanding how the brain processes a range of sensory data – visual, auditory and even tactile – as well as physical inference and scene understanding.
We envision TDW and PyImpact supporting research into human – and machine -- audio perception, that can lead to a better understanding of the computational principles underlying human audition. This understanding can, for example, ultimately help to create better assistive technology for the hearing-impaired.  We recognize that the diversity of “audio materials” used in PyImpact is not yet adequate to meet this longer-term goal, but we are actively addressing that and plan to increase the scope significantly.
We also believe the wide range of physics behaviors and interaction scenarios TDW supports will greatly benefit research into understanding how we as humans learn so much about the world, so rapidly and flexibly, given minimal input data. While we have made significant strides in the accuracy of physics behavior in TDW, TDW is not yet able to adequately support robotic simulation tasks. 
To support visual object recognition and image understanding we constantly strive to make TDW’s image generation as photoreal as possible using today’s real-time 3D technology. However, we are not yet at the level we would like to be.  We plan to continue improving our rendering and image generation capability, taking advantage of any relevant technology advances (e.g. real-time hardware-assisted ray tracing) while continuing to explore the relative importance of object variance, background variability and overall image quality to vision transfer results.

\section{Dataset Details}
\label{sec: dataset}

\subsection{TDW-image Dataset}
To generate images, the controller runs each model through two loops. The first loop captures camera and object positions, rotations, etc. Then, these cached positions are played back in the second loop to generate images. Image capture is divided this way because the first loop will "reject" a lot of images with poor composition; this rejection system doesn't require image data, and so sending image data would slow down the entire controller.

The controller relies on IdPassGrayscale data to determine whether an image has good composition. This data reduces the rendered frame of a segmentation color pass to a single pixel and returns the grayscale value of that pixel. To start the positional loop, the entire window is resized to 32 $\times$ 32 and render quality is set to minimal, in order to speed up the overall process. There are then two grayscale passes: One without occluding objects (by moving the camera and object high above the scene) and one with occluding scenery, but the exact same relative positions and rotations. The difference in grayscale must exceed 0.55 for the camera and object positions and rotations to be ``accepted". This data is then cached. In a third pass, the screen is resized back to 256 $\times$ 256, images and high-quality rendering are enabled, and the controller uses the cached positional/rotational data to iterate rapidly through the dataset.

\begin{figure}[http]
   \centering
   \includegraphics[width = 1.0\linewidth]{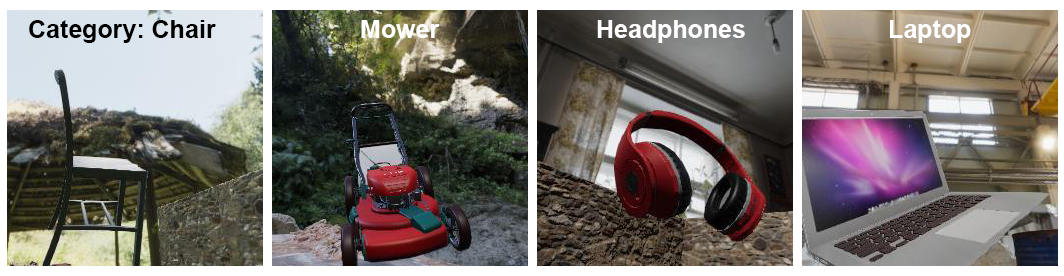}
    \vspace{-3mm}
   \caption{Examples from the TDW pre-training dataset, to be released as part of the TDW package.}
   \vspace{-3mm}
   \label{fig:tdw200}
\end{figure}

\subsection{Advanced Physical Prediction Benchmark}
Individual descriptions of each of the physics dataset scenarios as mentioned in the paper and shown in the Supplementary Material video. Note that additional scenarios are included here that were not mentioned in the paper; some are included in the video.

\noindent\textbf{Binary Collisions}
Randomly-selected "toys" are created with random physics values. A force of randomized magnitude is applied to one toy, aimed at another.

\noindent\textbf{Complex Collisions}
Multiple objects are dropped onto the floor from a height, with randomized starting positions and orientations.

\noindent\textbf{Object Occlusion}
Random "big" and "small" models are added. The small object is at random distance and angle from the big object. The camera is placed at a random distance and rotated such that the "big" model occludes the "small" model in some frames. Note -- not included in video.

\noindent\textbf{Object Permanence}
A ball rolls behind an occluding object and then reemerges. The occluder is randomly chosen from a list. The ball has a random starting distance, visual material, physics values, and initial force.

\noindent\textbf{Shadows}
A ball is added in a scene with a randomized lighting setup. The ball has a random initial position, force vector, physics values, and visual materials. The force vectors are such that the ball typically rolls through differently-lit areas, i.e. a bright spot to a shadowy spot.

\noindent\textbf{Stability}
A stack of 4-7 objects is created. The objects are all simple shapes with random colors. The stack is built according to a "stability" algorithm; some algorithms yield more balanced stacks than others. The stack falls down, or doesn't.

\noindent\textbf{Containment}
A small object is contained and rattles around in a larger object, such as a basket or bowl. The small object has random physics values. The bowl has random force vectors.

\noindent\textbf{Sliding/Rolling}
Objects are placed on a table. A random force is applied at a random point on the table. The objects slide or roll down.

\noindent\textbf{Bouncing}
Four "ramp" objects are placed randomly in a room. Two to six "toy" objects are added to the room in mid-air and given random physics values and force vectors, such that they will bounce around the scene. Note -- not included in video.

\noindent\textbf{Draping/Folding}
A cloth falls, 80 percent of the time onto another rigid body object. The cloth has random physics values.

\noindent\textbf{Dragging}
A rigid object is dragged or moved by pulling on a cloth under it. The cloth and the object have random physics values. The cloth is pulled in by a random force vector.

\noindent\textbf{Squishing}
Squishy objects deform and are restored to original shape depending on applied forces (e.g. squished when something else is on top of them or when they impact a barrier). Note -- not included in video.

\noindent\textbf{Submerging}
Objects sink or float in fluid. Values for viscosity, adhesion and cohesion vary by fluid type, as does the visual appearance of the fluid. Fluids represented in the video include water, chocolate, honey, oil and glycerin.

\section{Training a Learnable Intuitive Physics Simulator}
\label{dynamics}

\noindent\textbf{HRN Architecture.} We re-implemented the HRN architecture as published~\cite{mrowca2018flexible}, using the Tensorflow-2.1 library. To predict the future physical state, the HRN resolves physical constraints that particles connected in the hierarchical graph impose on each other. 
Graph convolutions are used to compute and propagate these effects.
Following \cite{Battaglia2016Interaction}, the HRN uses a \emph{pairwise graph convolution} with two basic building blocks: (1) A pairwise processing unit $\phi$ that takes the sender particle state $p_s$, the receiver particle state $p_r$ and their relation $r_{sr}$ as input and outputs the effect $e_{sr} \in \R^E$ of $p_s$ on $p_r$, and (2) a commutative aggregation operation $\Sigma$ which collects and computes the overall effect $e_r \in \R^E$. In our case this aggregation is a simple summation over all effects on $p_r$. Together these two building blocks form a convolution on graphs. 
The HRN has access to the Flex particle representation of each object, which is provided at every simulation step by the environment. From this particle representation, we construct a hierarchical particle relationship scene graph representation $G_H$. Graph nodes correspond to either particles or groupings of other nodes and are arranged in a hierarchy, whereas edges represent constraints between nodes. The HRN as the dynamics model takes a history of hierarchical graphs $G_H^{(t-T, t]}$ as input and predicts the future particle states $P^{t+1}$. The model first computes collision effects between particles ($\phi^W_C$), effects of external forces ($\phi^W_F$), and effects of past particles on current particles ($\phi^W_F$) using pairwise graph convolutions.  The effects are then propagated through the particle hierarchy using a hierarchical graph convolution module $\eta^W$. First effects are propagated from leaf to ancestor particles (L2A), then within siblings (WG), and finally from ancestors to descendants (A2D). Finally, the fully-connected module $\psi^W$ computes the next particle states $P^{t+1}$ from the summed effects and past particle states.

\noindent\textbf{Dynamic Recurrent HRN (DRHRN) for large environments.} Representing environment components (floor, walls) at the particle resolution of small objects is inefficient. Decreasing the resolution is problematic as small objects might miss environment interactions. Instead, we propose to initially model environment components as a sparse triangular mesh. At any given time, we compute each object particle's contact point with the environment by intersecting a ray originating from the particle in the direction of the mesh surface normal. If the contact point is closer than distance $d$, we spawn particles onto the triangle surface at the resolution of the small object. We dynamically add these particles to the graph $G_H^t$ and connect them to the object particle. Conversely, we delete environment nodes from the graph when objects move away from the environment. With this novel dynamic resolution method, which we call the Dynamic Recurrent HRN (DRHRN), we can efficiently represent large environments that can be modeled with TDW.  

DRHRN also builds on the original HRN by introducing an improved training loss and recurrent component. Specifically, the DRHRN loss predicts the position delta between current and next state $\Delta p = p^{t+1} - p^{t}$ using $L_2$ loss ($L_{Delta}$). To preserve object structure and shape, we additionally match the pairwise distance between predicted particle positions within each object $d_{ij} = ||p_i - p_j||$ to the ground truth particle distances via $L_2$ loss ($L_{Structure}$). The total loss is equal to the $\alpha$ weighted sum of both loss terms: $L = \alpha L_{Structure} + (1 - \alpha) L_{Delta}$.

Iterative physics prediction models accumulate errors exponentially. Naively trained one-step physics predictors only operate on ground truth input and do not see their own predictions as input during training, despite being tested via unrolling the model. To make DRHRN robust against its own prediction errors, we therefore train the model recurrently in a state-invariant way, i.e. without using a hidden state, as physical dynamics is state-free. The overall loss is then defined as the sum of all per time step losses.

\section{TDW Lighting Model}

\label{sec:lighting}
The lighting model for both interior and exterior environments utilizes a single primary light source that simulates the sun and provides direct lighting, affecting the casting of shadows. In most interior environments, additional point or spot lights are also used to simulate the light coming from lighting fixtures in the space. 

General environment (indirect) lighting comes from ``skyboxes” that utilize High Dynamic Range images (HDRI).  Skyboxes are conceptually similar to a planetarium projection, while HDRI images are a special type of photographic digital image that contain more information than a standard digital image. Photographed at real-world locations, they capture lighting information for a given latitude and hour of the day.  This technique is widely used in movie special-effects, when integrating live-action photography with CGI elements.

TDW's implementation of HDRI lighting automatically adjusts:
\begin{itemize}
\item The elevation of the ``sun" light source to match the time of day in the original image; this affects the length of shadows.
\item The intensity of the ``sun" light, to match the shadow strength in the original image.

\item The rotation angle of the ``sun" light, to match the direction shadows are pointing in the original image .
\end{itemize}

By rotating the HDRI image, we can realistically simulate different viewing positions, with corresponding changes in lighting, reflections and shadowing in the scene (see the Supplementary Material video for an example).

TDW currently provides over 100 HDRI images captured at various locations around the world and at different times of the day, from sunrise to sunset. These images are evenly divided between indoor and outdoor locations.




\section{Related Simulation Environments}
\label{sec:related}
Recently, several simulation platforms have been developed to support research into embodied AI, scene understanding, and physical inference. These include AI2-THOR\cite{kolve2017ai2}, HoME\cite{wu2018building}, VirtualHome\cite{puig2018virtualhome}, Habitat\cite{savva2019habitat},  Gibson\cite{xia2018gibson}, iGibson~\cite{xia2020interactive},
Sapien~\cite{xiang2020sapien} PyBullet~\cite{coumans2016pybullet}, MuJuCo~\cite{todorov2012mujoco}, and Deepmind Lab~\cite{beattie2016deepmind}. However none of them approach TDW's range of features and diversity of potential use cases. 

\noindent\textbf{Rendering and Scene Types.}
Research in computer graphics (CG) has developed extremely photorealistic rendering pipelines~\cite{kuhlo2010architectural}. 
However, the most advanced techniques (e.g. ray tracing), have yet to be fully integrated into real-time rendering engines. 
Some popular simulation platforms, including Deepmind Lab~\cite{beattie2016deepmind} and OpenAI Gym~\cite{brockman2016openai}, do not target realism in their rendering or physics and are better suited to prototyping than exploring realistic situations. Others use a variety of approaches for more realistic visual scene creation -- scanned from actual environments (Gibson, Habitat), artist-created (AI2-THOR) or using existing datasets such as SUNCG~\cite{song2016ssc} (HoME). However all are limited to the single paradigm of rooms in a building, populated by furniture, whereas TDW supports real-time near-photorealistic rendering of both indoor and outdoor environments. Only TDW allows users to create custom environments procedurally, as well as populate them with custom object configurations for specialized use-cases. For example, it is equally straightforward with TDW to arrange a living room full of furniture (see Fig. 1a-b), to generate photorealistic images of outdoor scenes (Fig. 1c) to train networks for transfer to real-world images, or to construct a ``Rube Goldberg” machine for physical inference experiments (Fig. 1h).

\noindent\textbf{Physical Dynamics.}
Several stand-alone physics engines are widely used in AI training, including PyBullet and MuJuCo which support a range of accurate and complex physical interactions. 
However, these engines do not generate high-quality images or audio output.
Conversely, platforms with real-world scanned environments, such as Gibson and Habitat, do not support free interaction with objects. 
HoME does not provide photorealistic rendering but does support rigid-body interactions with scene objects, using either simplified (but inaccurate) "box-collider" bounding-box approximations or the highly inefficient full object mesh. 
AI2-THOR provides better rendering than HoME or VirtualHome, with similar rigid-body physics to HoME. 
In contrast, TDW automatically computes convex hull colliders that provide mesh-level accuracy with box-collider-like performance (Fig. 2).
This fast-but-accurate high-res rigid body (denoted ``RF'' in Table 1) appears unique among integrated training platforms.
Also unique is TDW's support for complex non-rigid physics, based on the NVIDIA FLeX engine (Fig. 1d).
Taken together, TDW is substantially more full-featured for supporting future development in rapidly-expanding research areas such as learning scene dynamics for physical reasoning~\cite{yi2019clevrer, YeTian_physicsECCV2018} and model-predictive planning and control ~\cite{Battaglia2013Simulation,Mottaghi2016What,Fragkiadaki2016Learning,Battaglia2016Interaction,Chang2017compositional,Agrawal2016Learning,Shao2014Imagining,fire2016learning,pearl2009causality}. 

\noindent\textbf{Audio.}
As with CG, advanced work in computer simulation has developed powerful methods for physics-based sound synthesis~\cite{james2006precomputed} based on object material and object-environment interactions. In general, however, such physics-based audio synthesis has not been integrated into real-time simulation platforms. HoME and PyBullet are the only other platforms to provide audio output, generated by user-specified pre-placed sounds. TDW, on the other hand, implements a physics-based model to generate situational sounds from object-object interactions (Fig. 1h). TDW's PyImpact Python library computes impact sounds via modal synthesis with mode properties sampled from distributions conditioned upon properties of the sounding object~\cite{traer2019impacts}. The mode distributions were measured from recordings of impacts. The stochastic sound generation prevents overfitting to specific audio sequences. In human perceptual experiments, listeners could not distinguish our synthetic impact sounds from real impact sounds, and could accurately judge physical properties from the synthetic audio\cite{traer2019impacts}. For this reason, TDW is substantially more useful for multi-modal inference problems such as learning shape and material from sound~\cite{wangcomputational,zhang2017shape}.

\noindent\textbf{Interaction and API}
All the simulation platforms discussed so far require some form of API to control an agent, receive state of the world data or interact with scene objects. However not all support interaction with objects within that environment. Habitat focuses on navigation within indoor scenes, and its Python API is comparable to TDW’s but lacks capabilities for interaction with scene objects via physics (Fig. 1e), or multi-modal sound and visual rendering (Fig. 1h). 
VirtualHome, iGibson and AI2-THOR's interaction capabilities are closer to TDW's. In VirtualHome and AI2-THOR, interactions with objects are explicitly animated, not controlled by physics. TDW's API, with its multiple paradigms for true physics-based interaction with scene objects, provides a set of tools that enable the broadest range of use cases of any available simulation platform.

\section{System overview and API}
\label{sec:system}

\includegraphics[width=\linewidth]{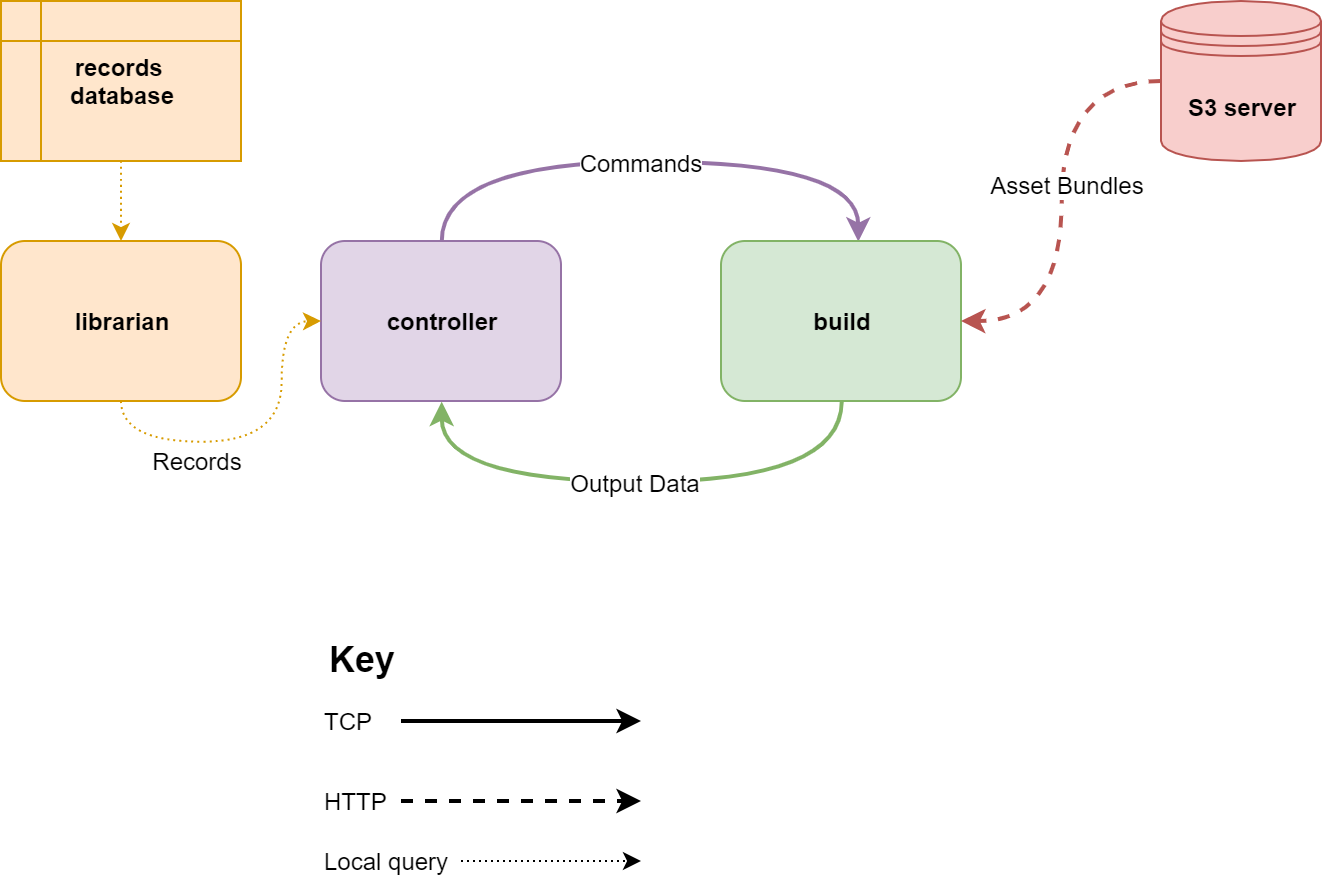}

\subsection{Core components}

\begin{enumerate}
    \item \textbf{The build} is the 3D environment application. It is available as a compiled executable.
    \item \textbf{The controller} is an external Python script created by the user, which communicates with the build.
    \item \textbf{The S3 server} is a remote server. It contains the  binary files of each model, material, etc. that can be added to the build at runtime.
    \item \textbf{The records databases} are a set of local .json metadata files with records corresponding to each asset bundle.
    \item A \textbf{librarian} is a Python wrapper class to easily query metadata in a records database file.
\end{enumerate}

\subsection{The simulation pattern}

\begin{enumerate}
    \item The controller communicates with the build by sending a list of \textbf{commands}.
    \item The build receives the list of serialized Commands, deserializes them, and executes them.
    \item The build advances 1 physics frame (simulation step).
    \item The build returns \textbf{output data} to the controller.
\end{enumerate}

Output data is always sent as a list, with the last element of the list being the frame number:

\texttt{[data, data, data, frame]}

\subsection{The controller}

All controllers are sub-classes of the \texttt{Controller} class. Controllers send and receive data via the \texttt{communicate} function:

\begin{verbatim}
from tdw.controller import Controller

c = Controller()

# resp will be a list with one element: [frame]
resp = c.communicate({"$type": "load_scene", "scene_name": "ProcGenScene"})
\end{verbatim}

Commands can be sent in lists of arbitrary length, allowing for arbitrarily complex instructions per frame.
The user must explicitly request any other output data:

\begin{verbatim}
from tdw.controller import Controller
from tdw.tdw_utils import TDWUtils
from tdw.librarian import ModelLibrarian
from tdw.output_data import OutputData, Bounds, Images

lib = ModelLibrarian("models_full.json")
# Get the record for the table.
table_record = lib.get_record("small_table_green_marble")

c = Controller()

table_id = 0

# 1. Load the scene.
# 2. Create an empty room (using a wrapper function)
# 3. Add the table.
# 4. Request Bounds data.
resp = c.communicate([{"$type": "load_scene",
                       "scene_name": "ProcGenScene"},
                      TDWUtils.create_empty_room(12, 12),
                      {"$type": "add_object",
                       "name": table_record.name,
                       "url": table_record.get_url(),
                       "scale_factor": table_record.scale_factor,
                       "position": {"x": 0, "y": 0, "z": 0},
                       "rotation": {"x": 0, "y": 0, "z": 0},
                       "category": table_record.wcategory,
                       "id": table_id},
                      {"$type": "send_bounds",
                       "frequency": "once"}])
\end{verbatim}

The \texttt{resp} object is a list of byte arrays that can be deserialized into output data:

\begin{verbatim}
# Get the top of the table.
top_y = 0
for r in resp[:-1]:
    r_id = OutputData.get_data_type_id(r)
    # Find the bounds data.
    if r_id == "boun":
        b = Bounds(r)
        # Find the table in the bounds data.
        for i in range(b.get_num()):
            if b.get_id(i) == table_id:
                top_y = b.get_top(i)
\end{verbatim}

The variable \texttt{top\_y} an be used to place an object on the table:

\begin{verbatim}
box_record = lib.get_record("iron_box")
box_id = 1
c.communicate({"$type": "add_object",
               "name": box_record.name,
               "url": box_record.get_url(),
               "scale_factor": box_record.scale_factor,
               "position": {"x": 0, "y": top_y, "z": 0},
               "rotation": {"x": 0, "y": 0, "z": 0},
               "category": box_record.wcategory,
               "id": 1})
\end{verbatim}

Then, an ``avatar" can be added to the scene. In this case, the avatar is a just a camera. The avatar can then send an image:

\begin{verbatim}
avatar_id = "a"
resp = c.communicate([{"$type": "create_avatar",
                       "type": "A_Img_Caps_Kinematic",
                       "avatar_id": avatar_id},
                      {"$type": "teleport_avatar_to",
                       "position": {"x": 1, "y": 2.5, "z": 2}},
                      {"$type": "look_at",
                       "avatar_id": avatar_id,
                       "object_id": box_id},
                      {"$type": "set_pass_masks",
                       "avatar_id": avatar_id,
                       "pass_masks": ["_img"]},
                      {"$type": "send_images",
                       "frequency": "once",
                       "avatar_id": avatar_id}])

# Get the image.
for r in resp[:-1]:
    r_id = OutputData.get_data_type_id(r)
    # Find the image data.
    if r_id == "imag":
        img = Images(r)
\end{verbatim}

This image is a numpy array that can be either saved to disk or fed directly into a ML system.Put together, the example code will create this image:

\includegraphics[width=0.6\linewidth]{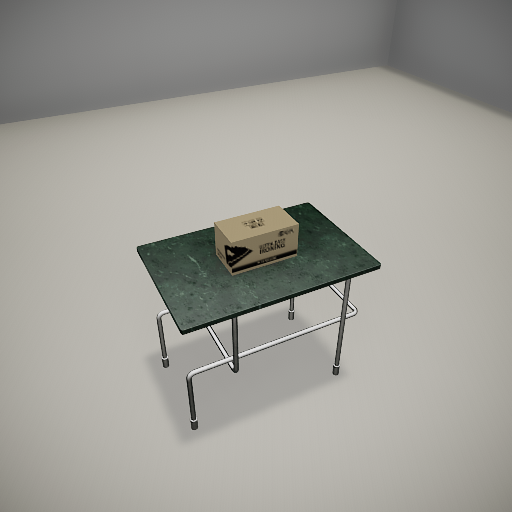}

\subsection{Benchmarks}

\textbf{CPU:}  Intel i7-7700K @4.2GHz \textbf{GPU:} NVIDIA GeForce GTX 1080

\begin{center}

\begin{tabular}{l|l|l|l}
    \hline
    \textbf{Benchmark} & \textbf{Quality} & \textbf{Size} & \textbf{FPS} \\
    \hline
    Object data & N/A & N/A & 850 \\
    \hline
    Images & low & 256x256 & 380 \\
    \hline
    Images & high & 256x256 & 168 \\
    \hline
\end{tabular}

\end{center}

\subsection{Command API Backend}

\subsubsection{Implementation Overview}

Every command in the Command API is a subclass of \texttt{Command}.

\begin{verbatim}
/// <summary>
/// Abstract class for a message sent from the controller to the build.
/// </summary>
public abstract class Command
{
    /// <summary>
    /// True if command is done.
    /// </summary>
    protected bool isDone = false;


    /// <summary>
    /// Do the action.
    /// </summary>
    public abstract void Do();


    /// <summary>
    /// Returns true if this command is done.
    /// </summary>
    public bool IsDone()
    {
        return isDone;
    }
}
\end{verbatim}

Every command must override \texttt{Command.Do()}. Because some commands require multiple frames to finish, they announce that they are ``done" via \texttt{Command.IsDone()}.

\begin{verbatim}
///<summary>
/// This is an example command.
/// </summary>
public class ExampleCommand : Command
{
    ///<summary>
    /// This integer will be output to the console.
    /// </summary>
    public int integer;
	
	
    public override void Do()
    {
        Debug.Log(integer);
        isDone = true;
    }
}
\end{verbatim}

\textbf{Commands are \textit{automatically} serialized and deserialized as JSON dictionaries} In a user-made controller script, \texttt{ExampleCommand} looks like this:

\begin{verbatim}
{"$type": "example_command", "integer": 15}
\end{verbatim}

If the user sends that JSON object from the controller, the build will deserialize it to an \texttt{ExampleCommand}-type object and call \texttt{ExampleCommand.Do()}, which will output \texttt{15} to the console.

\subsubsection{Type Inheritance}

The Command API relies heavily on type inheritance, which is handled automatically by the JSON converter. Accordingly, new commands can easily be created without affecting the rest of the API, and bugs affecting multiple commands are easy to identify and fix.

\begin{verbatim}
/// <summary>
/// Manipulate an object that is already in the scene.
/// </summary>
public abstract class ObjectCommand : Command
{
    /// <summary>
    /// The unique object ID.
    /// </summary>
    public int id;


    public override void Do()
    {
        DoObject(GetObject());
        isDone = true;
    }


    /// <summary>
    /// Apply command to the object.
    /// </summary>
    /// <param name="co">The model associated with the ID.</param>
    protected abstract void DoObject(CachedObject co);

    /// <summary>
    /// Returns a cached model, given the ID.
    /// </summary>
    protected CachedObject GetObject()
    {
        // Additional code here.
    }
}


/// <summary>
/// Set the object's rotation such that its forward directional vector points
/// towards another position.
/// </summary>
public class ObjectLookAtPosition : ObjectCommand
{
    /// <summary>
    /// The target position that the object will look at.
    /// </summary>
    public Vector3 position;


    protected override void DoObject(CachedObject co)
    {
        co.go.transform.LookAt(position);
    }
}
\end{verbatim}

The TDW backend includes a suite of auto-documentation scripts that scrape the \texttt{<summary>} comments to generate a markdown API page complete with example JSON per command, like this:

\includegraphics[width=\linewidth]{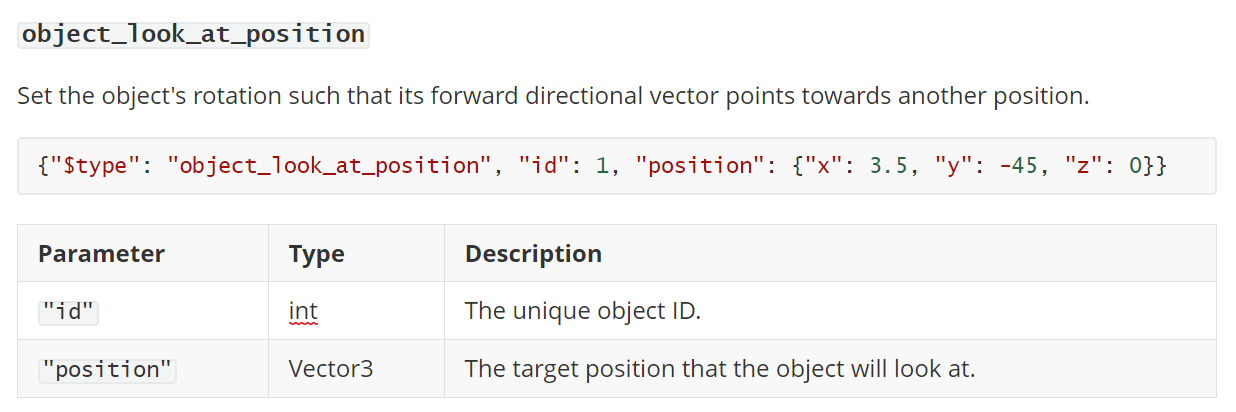}


\end{document}